\documentclass[sigconf]{acmart}

\usepackage{booktabs} 
\usepackage{lipsum}
\usepackage{multirow}
\usepackage{array}
\usepackage{courier}
\usepackage{mathtools}
\usepackage{caption}
\usepackage{float}
\usepackage{subcaption}
\usepackage{amsfonts}
\usepackage{balance}
\usepackage{tikz}
\usepackage{mathtools}

\begin{document}
\copyrightyear{2018}
\acmYear{2018}
\setcopyright{acmcopyright}
\acmConference[KDD '18]{The 24th ACM SIGKDD International Conference on Knowledge Discovery \& Data Mining}{August 19--23, 2018}{London, United Kingdom}
\acmBooktitle{KDD '18: The 24th ACM SIGKDD International Conference on Knowledge Discovery \& Data Mining, August 19--23, 2018, London, United Kingdom}
\acmPrice{15.00}
\acmDOI{10.1145/3219819.3220086}
\acmISBN{978-1-4503-5552-0/18/08}
\title{Multi-Pointer Co-Attention Networks for Recommendation}

\author{Yi Tay}
\affiliation{%
  \institution{Nanyang Technological University Singapore}
}
\email{ytay017@e.ntu.edu.sg}

\author{Luu Anh Tuan}
\affiliation{%
 \institution{Institute for Infocomm Research Singapore}
}
\email{at.luu@i2r.a-star.edu.sg}

\author{Siu Cheung Hui}
\affiliation{%
  \institution{Nanyang Technological University Singapore}
}
\email{asschui@ntu.edu.sg}

\begin{abstract}
Many recent state-of-the-art recommender systems such as D-ATT, TransNet and DeepCoNN exploit reviews for representation learning. This paper proposes a new
neural architecture for recommendation with reviews. Our model operates on a multi-hierarchical paradigm and is based on the intuition that
not all reviews are created equal, i.e., only a selected few are important. The importance, however,
should be dynamically inferred depending on the current target. To this end, we propose a review-by-review pointer-based learning scheme that extracts
important reviews from user and item reviews and subsequently matches them in a word-by-word fashion. This enables not only the most
informative reviews to be utilized for prediction but also a deeper word-level interaction. Our pointer-based method operates with
a gumbel-softmax based pointer mechanism that enables the incorporation of discrete vectors within differentiable neural architectures.
Our pointer mechanism is co-attentive in nature, learning pointers which are co-dependent on user-item relationships. Finally,
we propose a multi-pointer learning scheme that learns to combine multiple views of user-item interactions. We demonstrate the effectiveness of our proposed model via extensive experiments on \textbf{24} benchmark datasets from Amazon and Yelp. Empirical results show that
our approach significantly outperforms existing state-of-the-art models, with up to $19\%$ and $71\%$ relative improvement when compared to
TransNet and DeepCoNN respectively. We study the behavior of our multi-pointer learning mechanism, shedding light on `\textit{evidence aggregation}' patterns in review-based recommender systems.
 \end{abstract}


\maketitle

\section{Introduction}
On most e-commerce platforms today, the ability to write and share reviews is not only a central feature but is also a strongly encouraged act. Reviews are typically
informative, pooling an extensive wealth of knowledge for prospective customers. However, the extensive utility of reviews do not only end at this point. Reviews are also powerful in capturing preferences of authors, given the rich semantic textual information that cannot be conveyed via implicit interaction data or purchase logs. As such, there have been immense interest in collaborative filtering systems that exploit review information for making better recommendations \cite{DBLP:conf/kdd/DiaoQWSJW14,mcauley2013hidden,DBLP:conf/wsdm/ZhengNY17,cao2017cross,zhang2016collaborative,he2015trirank}.

Recent advances in deep learning has spurred on various innovative models that exploit reviews for recommendation \cite{DBLP:conf/wsdm/ZhengNY17,catherine2017transnets,Seo:2017:ICN:3109859.3109890}. The intuition is simple yet powerful, i.e., each user is represented as all reviews he (she) has written and an item is represented by all reviews that was written for it. All reviews are concatenated to form a single user (item) document. Subsequently, a convolutional encoder is employed to learn a single latent representation for the user (item). User and item embeddings are then matched using a parameterized function such as Factorization Machines \cite{DBLP:conf/icdm/Rendle10}. This has shown to be highly performant \cite{Seo:2017:ICN:3109859.3109890}, outperforming a wide range of traditionally strong baselines such as Matrix Factorization (MF). Models such as DeepCoNN \cite{DBLP:conf/wsdm/ZhengNY17}, TransNets \cite{catherine2017transnets} and D-ATT \cite{Seo:2017:ICN:3109859.3109890} are recent state-of-the-art models that are heavily grounded in this paradigm.

Intuitively, this modeling paradigm leaves a lot to be desired. Firstly, the naive concatenation of reviews into a single document is unnatural, ad-hoc and noisy. In this formulation, reviews are treated indiscriminatingly irregardless of whether they are important or not. A user's bad review about a coffee shop should be mostly irrelevant when deciding if a Spa is a good match. Secondly, user and item representations are static irregardless of the target match. For example, when deciding if a coffee shop is a good match for a user, the user's past reviews about other coffee shops (and eateries) should be highly relevant. Conversely, reviews about spas and gyms should not count. Hence, a certain level of dynamism is necessary. Finally, the only accessible interaction between user and item is through a fixed dimensional representation, which is learned via excessive compression of large user-item review banks into low-dimensional vector representations. For richer modeling of user and item reviews, deeper and highly accessible interaction interfaces between user-item pairs should be mandatory.

Recall that reviews were fundamentally written independently, at different times, and for different products (or by different people). There should be no reason to squash everything into one long document if they can be modeled independently and then combined later. More importantly, a user may write hundreds of reviews over an extended period of time while an item may effortlessly receive a thousand of reviews. As such, existing modeling paradigms will eventually hit a dead-end. Overall, this work proposes four major overhauls have to be made to existing models, i.e., (1) reviews should be modeled independently, (2) not all reviews are equally important and should be weighted differently, (3) the importance of each review should be dynamic and dependent on the target match and finally, (4) user and item reviews should interact not only through compressed vector representations but also at a deeper granularity, i.e., word-level.

To this end, we propose a \textit{Multi-Pointer Co-Attention Network} (MPCN), a novel deep learning architecture that elegantly satisfies our key desiderata. Our model is multi-hierarchical in nature, i.e., each user is represented by $n$ reviews of $\ell$ words each. Subsequently, all user and item reviews (for this particular instance pair) are matched to determine the most informative reviews. In order to do so, we design a novel pointer-based co-attention mechanism. The pointer mechanism extracts the named reviews for direct review-to-review matching. At this stage, another co-attention mechanism learns a fixed dimensional representation, by modeling the word-level interaction between these matched reviews. This forms the crux of our \textit{review-by-review} modeling paradigm. Finally, we introduce a multi-pointer learning scheme that can be executed an arbitrary $k$ times, extracting multiple multi-hierarchical interactions between user and item reviews.

\subsection{Our Contributions}
In summary, the prime contributions of this paper are as follows:
\begin{itemize}
\item We propose a state-of-the-art neural model for recommendation with reviews. Our proposed model exploits a novel
pointer-based learning scheme. This enables not only noise-free but also deep word-level interaction between user and item.
\item We conduct experiments on $\textbf{24}$ benchmark datasets. Our proposed MPCN model outperforms all state-of-the-art baselines by a significant margin across all datasets. Our compared baselines are highly competitive, encompassing not only review-based models but also state-of-the-art interaction-only models. We outperform models such as Neural Matrix Factorization (NeuMF) \cite{He:2017:NCF:3038912.3052569}, DeepCoNN \cite{DBLP:conf/wsdm/ZhengNY17}, D-ATT \cite{Seo:2017:ICN:3109859.3109890} and TransNet \cite{catherine2017transnets}. Performance improvement over DeepCoNN, TransNets and D-ATT are up to $71\%$, $19\%$ and $5\%$ respectively.
\item We investigate the inner workings of our proposed model and provide insight about how MPCN works under the hood. Additionally, analyzing the behavior of our pointer mechanism allows us to better understand the nature of the problem. Through analysis of our pointer mechanism, we show that different problem domains have different patterns of \textit{`evidence aggregation'}, which might require different treatment and special care.
\end{itemize}

\section{Related Work}
In this section, we identify and review several key areas which are highly relevant to our work.
\subsection{Review-based Recommendation}
The utility of exploiting reviews for recommendations have been extensively discussed and justified in many works \cite{DBLP:conf/wsdm/ZhengNY17,catherine2017transnets,Kim:2016:CMF:2959100.2959165,mcauley2013hidden,Seo:2017:ICN:3109859.3109890}. This not only enables a mitigation of cold-start issues but also provides a richer semantic modeling of user and item characteristics. While relatively earlier works have mainly concentrated efforts on topic modeling and language modeling approaches \cite{mcauley2013hidden,DBLP:conf/kdd/DiaoQWSJW14,wang2011collaborative}, the recent shift towards deep learning models is prominent. The advantages of neural architectures are clear, i.e., not only do these models dispense with laborious feature engineering altogether, they are often highly competitive. In many recent works, Convolutional neural networks (CNN) act as automatic feature extractors, encoding a user (item) into a low-dimensional vector representation. User and item embeddings are then compared with a matching function.

An earlier neural model, the Deep Co-operative Neural Networks (DeepCoNN) \cite{DBLP:conf/wsdm/ZhengNY17} represents a user as all the reviews that he (she) has written. Likewise, an item is represented as all the reviews ever written (by other users) for it. User and item documents are then encoded with CNNs and passed into a Factorization Machine (FM) \cite{DBLP:conf/icdm/Rendle10} for matching. It was later argued that DeepCoNN's competitive performance exploits the fact that test reviews were leaked (into the training set) \cite{catherine2017transnets}. As such, this reduces the recommendation problem to resemble a noisy adaptation of standard document-level sentiment analysis. To this end, \cite{catherine2017transnets} proposed TransNets, augmenting a DeepCoNN-like neural network with an additional multi-task learning scheme. More specifically, it learns to transform the penultimate hidden layer of DeepCoNN into a CNN-encoded representation of the test review. This signal was found to be useful, improving the performance on multiple benchmarks.

DeepCoNN and TransNet are relatively simple model architectures. Another recently proposed model, the Dual Attention CNN model (D-ATT) \cite{Seo:2017:ICN:3109859.3109890} proposed augmenting CNNs with neural attention. The key idea of neural attention \cite{bahdanau2014neural} is to emphasize important segments of documents by weighting each word by a learned \textit{attention vector}. The final representation comprises a weighted linear combination of all input embeddings. Two variants of attention mechanism are proposed, i.e., local and global, both modeling different views of user-item review documents. However, these models are not without flaws. As mentioned, these models follow the same paradigm of representing user and item as a giant concatenated document of all their reviews which suffers inherent drawbacks such as (1) noise, (2) lack of dynamic target-dependent and (3) lack of interaction interfaces.

\subsection{Text Matching and Co-Attentional Models}
Our work is closely related to the problem domain of sequence pair modeling. A wide spectrum of models have been proposed for modeling relationships between two sequences, e.g., question-answer \cite{DBLP:journals/corr/SantosTXZ16}, premise-hypothesis \cite{rocktaschel2015reasoning,DBLP:conf/emnlp/ParikhT0U16,1801.00102} which are very similar to the user-item modeling problem at hand. In these fields, learning representations without fine-grained interaction modeling, i.e., absence of interaction interfaces in DeepCoNN, is known to be far outperformed by new models which utilize co-attentional mechanisms \cite{DBLP:journals/corr/SantosTXZ16,DBLP:conf/aaai/ZhangLSW17}. Co-attentions learn pairwise attention between two sequences \cite{DBLP:journals/corr/XiongZS16} (or modalities \cite{lu2016hierarchical}), enabling pair-aware attention weights to be learned.
\subsection{Recent Advances in Deep Learning}
Notably, our network solely relies on attention mechanisms, and showcases the potential neural architectures that do not use convolutional and recurrent layers. This is inspired by the Transformer \cite{vaswani2017attention} architecture which uses multi-headed attention, concatenating outputs of each attention call. Consequently, our proposed model does not use any recurrent or convolution layers, and solely relies on attention.

Additionally, our work is characterized by the usage of pointers, which have been popularized by both Pointer Networks \cite{vinyals2015pointer}. These networks learn to predict an output token which exists in the sequence itself, i.e., pointing to a token. The usage of pointers is primarily motivated for discrete problems and has also been widely adopted for answer span prediction in question-answering \cite{wang2016machine}. In these models, pointers are commonly applied at the last layer and have no issues since the model optimizes a loss function such as the cross entropy loss. However, our model requires the usage of pointers \textit{within} the network (between layers). As such, a form of hard attention is required. Due to the non-differentiability of hard attention, it has been much less popular than the standard soft attention. In order to generate discrete pointers, we utilize the recent gumbel-softmax \cite{jang2016categorical} trick which is capable of learning one-hot encoded vectors. Notably, the gumbel-softmax was recently adapted for automatic compositional learning of Gumbel TreeLSTMs \cite{choi2017unsupervised}, in which the gumbel-softmax is exploited to adaptively learn to make merging decisions.

\subsection{Deep Learning for Recommendation}
Factorization-based models \cite{mnih2008probabilistic,DBLP:conf/icdm/Rendle10,DBLP:conf/sigir/HeZKC16} were popular standard machine learning baselines for interaction-based recommendation. Today, we see a shift into deep learning, in which neural models are claiming state-of-the-art performance \cite{He:2017:NCF:3038912.3052569,zhang2018neurec,Tay:2018:LRM:3178876.3186154,DBLP:conf/ijcai/XiaoY0ZWC17}. He et al. \cite{He:2017:NCF:3038912.3052569} proposed a neural framework that combines a generalized matrix factorization formulation with multi-layered perceptrons. He and Chua \cite{he2017neural} proposed a neural adaptation of factorization machines. \cite{DBLP:conf/wsdm/WuABSJ17} proposed recurrent models for sequential modeling of interaction data. Tay et al. \cite{Tay:2018:LRM:3178876.3186154} proposed a translation-based framework that exploits neural attention for modeling user-item relations. Li et al. \cite{li2017neural} proposed a encoder-decoder based framework for both rating prediction and generating tips. A separate class of deep models based on auto-encoders \cite{sedhain2015autorec,Li:2017:CVA:3097983.3098077,wang2015collaborative} has also been proposed for recommendation tasks.

\section{Our Proposed Model}{}
In this section, we present a layer-by-layer description of our proposed model. The overall model architecture is illustrated in Figure \ref{fig:architecture}.

\begin{figure}[ht]
\centering
  \includegraphics[width=0.96\linewidth]{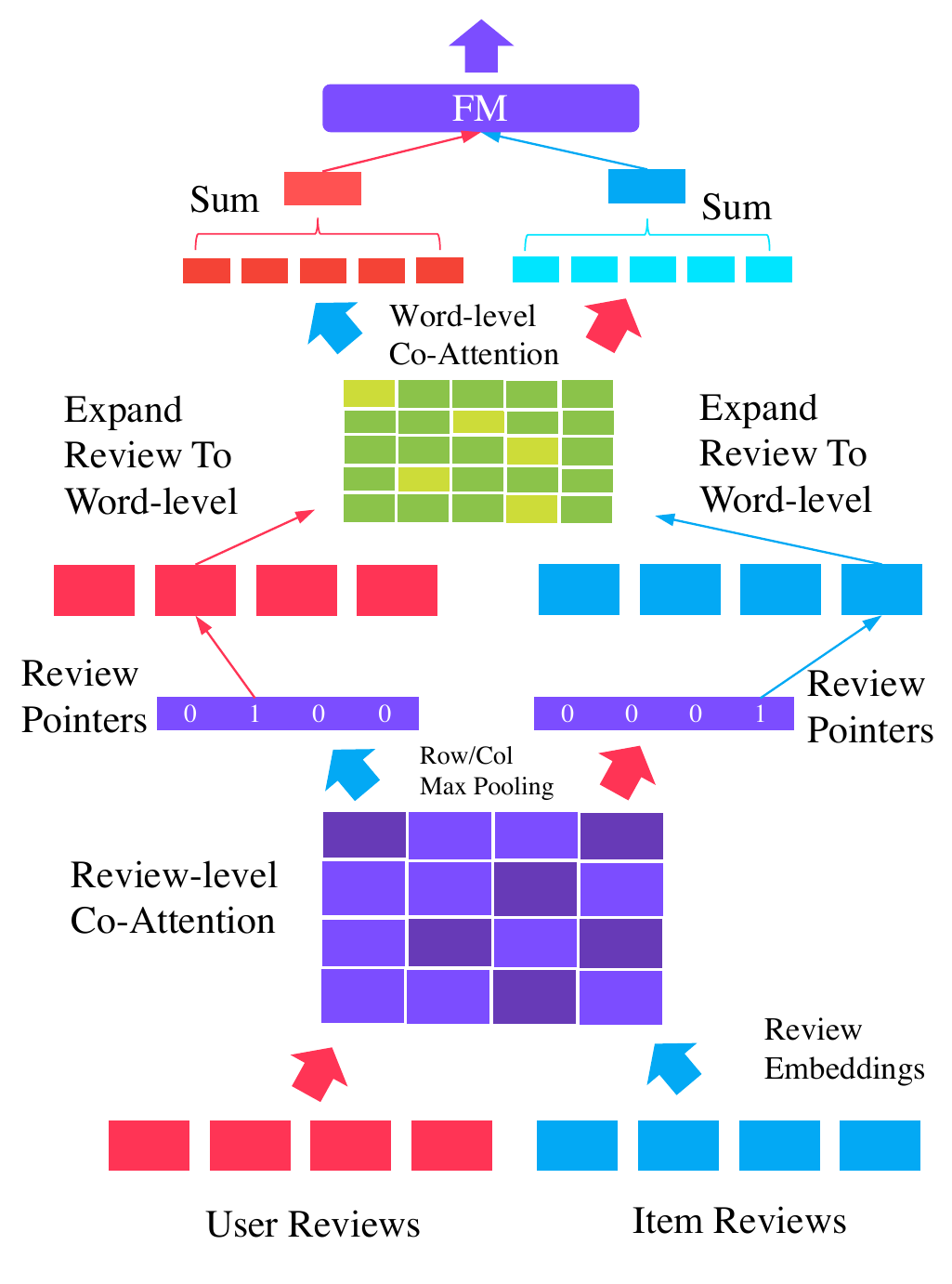}
  \caption{Illustration of proposed model architecture for Pointer-based Learning (\textit{Best viewed in color}). This example illustrates a one pointer example. Review gating mechanism and multi-pointer learning is omitted for clarity.}
  \label{fig:architecture}
\end{figure}

\subsection{Input Encoding}
Our model accepts two input sequences, $a$ (user) and $b$ (item). Each input sequence
is a list of reviews $\{r_1, r_2 \cdots r_{\ell_{d}} \}$ where $\ell_{d}$ is the maximum
number of reviews.
\subsubsection{Embedding Layer}
 Each review is a sequence of $\ell_{w}$ words which are
represented as one-hot encoded vectors. For $a$ and $b$, we pass
all words into an embedding matrix $\textbf{W}^{d \times |V|}$ where $V$ is the vocabulary,
retrieving a $d$ dimensional vector for each word. Our network is hierarchical in nature, i.e.,
instead of representing all user (or item) reviews as one long document, we represent
each user (or item) as a sequence of reviews. Each review is then hierarchically constructed
from a sequence of words.
\subsubsection{Review Gating Mechanism}
Each review is represented as a sum of its constituent word embeddings to form
the vector $x \in \mathbb{R}^{d}$. Intuitively, not all reviews that a user writes and not all reviews written for a product is important. We design
a first-stage filter, i.e., a review gating mechanism that accepts each review as an input and controls how much information passes through to the next level. Given the input $x \in \mathbb{R}^{\ell_{r} \times d}$, which represents either $a$ or $b$.
\begin{align}
\bar{x_{i}} = \sigma(\textbf{W}_{g}x_i) + \textbf{b}_{g} \odot tanh(\textbf{W}_{u}x_i + b_{u})
\end{align}
where $\odot$ is the Hadamard product and $\sigma$ is the sigmoid activation function. $x_i$ is the i-th review of sequence $x$. $\textbf{W}_{g}, \textbf{W}_{u} \in \mathbb{R}^{d \times d}$ and $b_g, b_u \in \mathbb{R}^{n}$ are parameters of this layer. While the purpose of the co-attention layer is to extract important reviews, we hypothesize that applying a pre-filter (gating mechanism) helps improve performance on certain datasets.

\subsection{Review-level Co-Attention}
In this layer, the aim is to select the most informative review from the review bank of each user
and item respectively.
\subsubsection{Affinity Matrix}
Given a list of review embeddings from each user ($a \in \mathbb{R}^{\ell_{r} \times d}$) and item ($b \in \mathbb{R}^{\ell_{r} \times d}$) banks, we
calculate an affinity matrix between them. This is described as follows:
\begin{align}
s_{ij} = F(a_{i})^{\top}\textbf{M} \:F(b_{j})
\label{affinty}
\end{align}
where $\textbf{M}^{d \times d}$ and $S \in \mathbb{R}^{\ell_r \times \ell_r}$. $F(.)$ is a feed-forward neural network function with $l$ layers. In practice, $l$ is tuned amongst $[0,2]$ where $l=0$ reverts Equation (\ref{affinty}) to the bilinear form.

\subsubsection{Pooling Function}
By taking the row and column wise maximum of the matrix $s$ and using them to weight the original list of reviews $a$ and $b$, we are able to derive the standard co-attention mechanism. This is described as follows:
\begin{align}
a' = (G(\max_{col}(s)))^{\top} a \: \: \: \text{and}\: \: \: \: \: b' = (G(\max_{row}(s)))^{\top} b
\label{ptr}
\end{align}
There are different choices for the pooling operation. Max pooling is used here because, intuitively it selects the review which has the maximum influence (or affinity) with all reviews from its partner. This type of co-attention is known to be extractive, characterized by its usage of max pooling.

Note that we apply the function $G(.)$ to $\max_{col}(s)$ and $\max_{row}(s)$. In most applications, $G(.)$ would correspond to the standard softmax function, which converts the input vector into a probability distribution. The vectors $a', b'$ would then be the \textit{co-attentional vector representations}. However, in our case, we desire further operations on the selected reviews and therefore do not make use of these representations. Instead, $G(.)$ has to return a one-hot encoded vector, pointing to the selected reviews which forms the real objective behind this co-attentional layer. However, the Softmax function returns a continuous vector, which is unsuitable for our use-case. The usage of discrete vectors in neural architectures is known to be difficult, as the $\arg \max$ operation is non-differentiable. Hence, we leverage a recent advance, the Gumbel-Max trick, for learning to point. The next section describes this mechanism.

\subsection{Review Pointers}
We leverage a recent advance, the Gumbel-Softmax \cite{jang2016categorical}, for incorporating discrete random variables in the network.

\subsubsection{Gumbel-Max}
In this section, we describe Gumbel-Max \cite{maddison2014sampling}, which facilitates the key mechanism of our MPCN model. Gumbel-Max enables discrete random variables (e.g., one-hot vectors) to be utilized within an end-to-end neural network architecture. Consider a $k$-dimensional categorical distribution where class probabilities $p_1, \cdots p_{k}$ are defined in terms of unnormalized log probabilities $\pi_{1}, \cdots \pi_{k}$:
\begin{align}
p_i = \frac{\exp(\log(\pi_i))}{\sum^{k}_{j=1} \exp (\log(\pi_j))}
\end{align}
A one-hot sample $z=(z_1, \cdots z_k) \in \mathbb{R}^{k}$ from the distribution can be drawn by using the following:
\begin{equation}
z_{i} =
\begin{cases}
1 \:, i = \arg\max_j (\log(\pi_j) + g_j) \\
0 \: , \text{otherwise}
\end{cases}
 \end{equation}
 \begin{align}
g_i = - \log (-\log (u_i)) \:\: \:\:\:\:\: u_i \sim Uniform(0,1)
\end{align}
where $g_i$ is the \textit{Gumbel noise} which perturbs each $\log(\pi_i)$ term such that the $\arg \max$ operation is equivalent
to drawing a sample weighted by $p_i, \cdots p_k$.

\subsubsection{Straight-Through Gumbel-Softmax}
In the Gumbel-Softmax, the key difference is that the $\arg\max$ function is replaced by the differentiable softmax function which is described as follows:
\begin{align}
y_i = \frac{\exp(\frac{\log(\pi_i) + g_i}{\tau})}{\sum^{k}_{j=1} \exp(\frac{\log(\pi_j) + g_i}{\tau})}
\end{align}
where $\tau$, the temperature parameter, controls the extend of how much the output approaches a one hot vector. More concretely, as $\tau$ approaches zero, the output of the Gumbel-Softmax distribution becomes cold, i.e., becomes closer to a one-hot vector. In the straight-through (ST) adaptation, the forward-pass is discretized by converting the vector output to a one-hot vector via $\arg\max$:
\begin{equation}
y_{i} =
\begin{cases}
1 \:, i = \arg\max_j (y_j) \\
0 \: , \text{otherwise}\\
\end{cases}
 \end{equation}
However, the backward pass maintains the flow of continuous gradients which allows the model to be trained end-to-end. This is useful as we only want important reviews to be selected (i.e., hard selection) to be considered in computation of the loss function. Notably, alternatives such as REINFORCE \cite{williams1992simple} exist. However, it is known to suffer from high variance and slow convergence \cite{jang2016categorical}.
\subsubsection{Learning to Point}
In order to compute a review pointer (for user and item), we set $G(.)$ in Equation (\ref{ptr}) to use the Gumbel Softmax. However, since we are interested
only in the pointer (to be used in subsequent layers), the pointer is then calculated as:
\begin{align}
p_{a} = (Gumbel(\max_{col}(s)))\: \:\: \text{and}  \: \: \: \: p_{b} = (Gumbel(\max_{row}(s)))
\label{ptr2}
\end{align}
By applying $p_{a}$ to $a$, we select the $p_{a}-th$ review of user $a$ and $p_{b}-th$ review of item $b$. The selected reviews are then passed into the next layer where rich interactions are extracted between these reviews.
\subsection{Word-level Co-Attention}
The review-level co-attention smooths over word information as it compresses each review into a single embedding. However, the design of the model allows the most informative reviews to be extracted by the use of pointers. These reviews can then be compared and modeled at word-level. This allows a user-item comparison of finer granularity which facilitates richer interactions as compared to simply composing the two review embeddings. Let $\bar{a},\bar{b}$ be the selected reviews using the pointer learning scheme. Similar to the review-level co-attention, we compute a similarity matrix between $\bar{a}$ and $\bar{b}$. The key difference is that the affinity matrix is computed word-by-word and not review-by-review.
\begin{align}
w_{ij} = F(\bar{a}_{i})^{\top}\textbf{M}_w \:F(\bar{b}_{j})
\end{align}
where $\textbf{M}_{w} \in \mathbb{R}^{d \times d}$ and $w \in \mathbb{R}^{\ell_{w} \times \ell_{w}}$. Next, to compute the \textit{co-attentional} representation of reviews $\bar{a}, \bar{b}$, we take the mean pooling.
\begin{align}
\bar{a}' = (S(avg_{col}(w)))^{\top} \bar{a} \: \: \: \text{and}\: \: \: \: \bar{b}' = (S(avg_{row}(w)))^{\top} \bar{b}
\label{word_coattention}
\end{align}
where $S(.)$ is the standard Softmax function and $F(.)$ is a standard feed-forward neural network with $l$ layers. The rationale for using the average pooling operator here is also intuitive. At the review-level, a large maximum affinity score of a review with respect to all \textit{opposite} reviews (even when a low average affinity socre) warrants it being extracted, i.e., a strong signal needs to be further investigated. However, at a word-level, max-pooling may be biased towards identical words and may have a great inclination to act as a word matching operator. Hence, we want to maintain a stable co-attentive extractor. Our early empirical experiments also justify this design. Finally $\bar{a}'$ and $\bar{b}'$ are the output representations.

Note that, implementation of co-attention layers (review-level and word-level) is equivalent to only two simple \textsc{matmul} operations (in Tensorflow). As such, scalability is not really a concern in our approach since this is quite efficiently optimized on GPUs.

\subsection{Multi-Pointer Learning}
While our objective is to eliminate noisy reviews by the usage of hard pointers, there might be insufficient information if we point to only a single pair of reviews. Hence, we devise a multi-pointer composition mechanism. The key idea is to use multiple pointers where the number of pointers $n_p$ is a user-defined hyperparameter. Our model runs the Review-level Co-Attention $n_p$ times, with each generating a unique pointer. Each of the $n_p$ review pairs is then modeled with the Word-level Co-Attention mechanism. The overall output is a list of vectors $\{ \bar{a}'_1, \cdots \bar{a}'_{n_p} \}$ and $\{ \bar{b}'_1, \cdots \bar{b}'_{n_p} \}$.  Additionally, we also found it useful to include the sum embedding of all words belonging to the user (item). This mainly helps in robustness, in case where user and item do not have any matching signals that were found by our pointer mechanism. We explore three ways to compose these embeddings.
\begin{itemize}
	\item \textbf{Concatenation} - All pointer outputs are concatenated, e.g., $[\bar{a}'_1; \cdots \bar{a}'_{n_p}]$. This has implications to the subsequent layers, especially if $n_p$ is large. Hence we consider the next two alternatives.
\item \textbf{Additive Composition} - All pointer outputs are summed, e.g., $sum(\bar{a}'_1, \cdots \bar{a}'_{n_p})$.

\item \textbf{Neural Network} - All pointer outputs are concatenated and passed through a single non-linear layer with ReLU ($\sigma_r$) activations. e.g., $\sigma_r(W([\bar{a}'_1; \cdots \bar{a}'_{n_p}]) +b)$ which maps the concatenated vector into a $d$ dimensional vector.
\end{itemize}
Note that this is applied to $\bar{b}'$ as well but omitted for brevity. Let the output of this layer be $a_{f}$ and $b_{f}$. In our experiments, we tune amongst the three above-mentioned schemes. More details are provided in the ablation study.

\begin{figure}[ht]
\centering
  \includegraphics[width=1.0\linewidth]{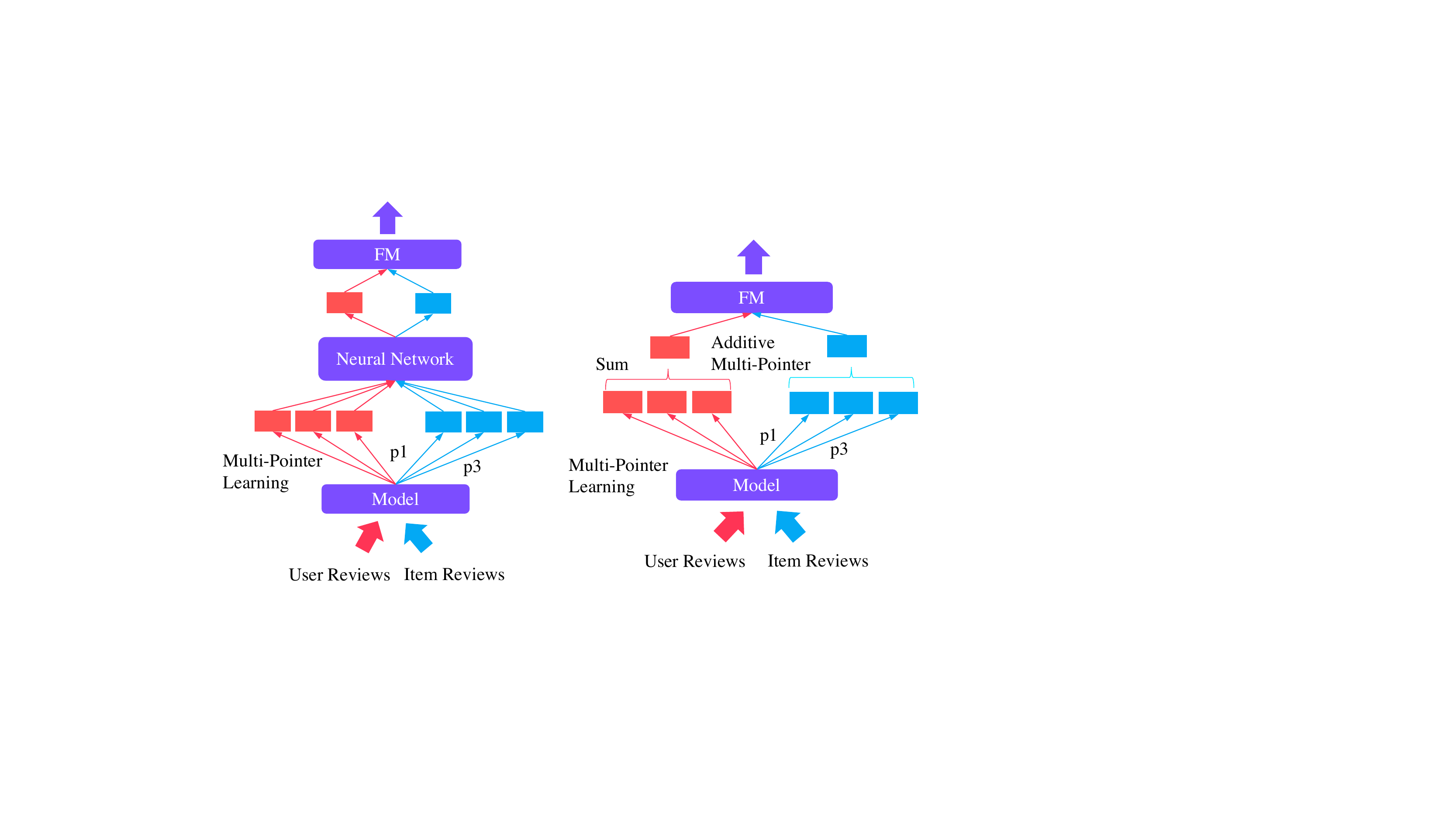}
  \caption{Illustration of Neural Network (\textit{left}) and Additive (\textit{right}) based Multi-Pointer Learning.}
  \label{fig:architecture}
\end{figure}

\subsection{Prediction Layer}
This layer accepts $a_{f}, b_{f}$ as an input. The concatenation of $[a_f; b_f]$ is passed into a factorization machine (FM) \cite{DBLP:conf/icdm/Rendle10}. FM accepts a real-valued feature vector and models the pairwise interactions between features using factorized parameters. The FM function is defined as follows:
 \begin{align}
F(x) & = w_{0} + \sum^{n}_{i=1} w_i \: x_i  +  \sum^{n}_{i=1} \sum^{n}_{j=i+1} \langle v_i, v_j \rangle \: x_i \: x_j
\end{align}
where $x \in \mathbb{R}^{k}$ is a real-valued input feature vector. $\langle .,. \rangle$ is the dot product. The parameters $\{v_{1} \dots v_{n} \}$ are factorized parameters (vectors of $v \in \mathbb{R}^{k}$) used to model pairwise interactions $(x_i, x_j)$. $w_0$ is the global bias and $\sum^{n}_{i=1} w_i \: x_i $ represents a linear regression component. The output of $F(x)$ is a scalar, representing the strength of the user-item interaction. The network is trained end-to-end by minimizing the standard mean squared error loss following \cite{Seo:2017:ICN:3109859.3109890}.

\section{Empirical Evaluation}
In this section, we present our experimental setup and empirical evaluation. Our experiments are designed to answer the following
research questions (RQs):
\begin{enumerate}
\item \textbf{RQ1} - Does our proposed approach outperform state-of-the-art models such as D-ATT and DeepCoNN? How much is the relative improvement?
\item \textbf{RQ2} - What are the impacts of some of the design / architectural choices of MPCN?
\item \textbf{RQ3} - What are the effects of the key hyperparameters of our model (e.g., number of pointers, etc.) on model performance?

\item \textbf{RQ4} - Are we able to derive any insight about how MPCN works by analyzing the behavior of the pointer layer?

\end{enumerate}
\subsection{Datasets}
We utilize datasets from two different sources which are described as follows:
\begin{enumerate}
\item \textbf{Yelp Dataset Challenge} - Yelp is an online review platform for businesses such as restaurants, bars, spas, etc. We use the dataset from the latest challenge\footnote{\url{https://www.yelp.com/dataset/challenge}}.
\item \textbf{Amazon Product Reviews} - Amazon is a well-known E-commerce platform. Users are able to write reviews for products they have purchased. We use \textbf{23} datasets from the Amazon Product Review corpus\footnote{\url{http://jmcauley.ucsd.edu/data/amazon/}} \cite{mcauley2015image,he2016ups}.
\end{enumerate}
In total, we provide model comparisons over $\textbf{24}$ benchmark datasets. For all datasets, we split interactions into training, development and testing sets. We utilize a time-based split, i.e., the last item of each user is added to the test set while the penultimate is used for development. For Amazon, the datasets are preprocessed in a 5-core fashion (i.e., each user and item have at least 5 reviews to be included). Since the datasets can be found in the official webpage, we do not restate their statistics to save space. For Yelp, we use a 20-core setting, providing a comparison on a denser dataset. We tokenize the reviews using NLTK and retain words that appear at least $10$ times in the vocabulary. We would like to emphasize that, when building user and item representations using their respective reviews, all reviews belonging to interactions from the test and development sets \textbf{were not included}. This is to prevent the problem from reverting to a sentiment analysis task, albeit noisier \cite{catherine2017transnets}.
\begin{table*}[htbp]
  \centering
    \begin{tabular}{|l|cccc|cccc|ccc|}
    \hline
      & \multicolumn{4}{c|}{Interaction-based} & \multicolumn{4}{c|}{Review-based} & \multicolumn{3}{c|}{Improvement $(\%)$} \\
    \hline
    Dataset & \multicolumn{1}{c}{MF} & \multicolumn{1}{c}{FM} & \multicolumn{1}{c}{MLP} & \multicolumn{1}{c|}{NEUMF} & \multicolumn{1}{c}{D-CON} & \multicolumn{1}{c}{TNET} & \multicolumn{1}{c}{D-ATT} & \multicolumn{1}{c|}{MPCN} & \multicolumn{1}{c}{$\Delta_{DC}$} & \multicolumn{1}{c}{$\Delta_{TN}$}
    & \multicolumn{1}{c|}{$\Delta_{DA}$} \\
    \hline
    Yelp17 & 1.735 & 1.726 & 1.727 & 1.691 & 1.385 & 1.363 & 1.428 & \textbf{1.349} & 2.7   & 1.0   & 5.9 \\
    Instant Video & 2.769 & 1.331 & 1.532 & 1.451 & 1.285 & 1.007 & 1.004 & \textbf{0.997} & 28.9  & 1.0   & 0.7 \\
    Instruments & 6.720 & 1.166 & 1.081 & 1.189 & 1.483 & 1.100 & 0.964 & \textbf{0.923} & 60.7  & 19.2  & 4.4 \\
    Digital Music & 1.956 & 1.382 & 1.361 & 1.332 & 1.202 & 1.004 & 1.000 & \textbf{0.970} & 23.9  & 3.5   & 3.1 \\
    Baby  & 1.755 & 1.614 & 1.585 & 1.598 & 1.440 & 1.338 & 1.325 & \textbf{1.304} & 10.4  & 2.6   & 1.6 \\
    Patio / Lawn & 2.813 & 1.311 & 1.279 & 1.251 & 1.534 & 1.123 & 1.037 & \textbf{1.011} & 51.7  & 11.1  & 2.6 \\
    Gourmet Food  & 1.537 & 1.348 & 1.442 & 1.464 & 1.199 & 1.129 & 1.143 & \textbf{1.125} & 6.6   & 0.4   & 1.6 \\
    Automotive & 5.080 & 1.599 & 1.071 & 1.013 & 1.130 & 0.946 & 0.881 & \textbf{0.861} & 31.2  & 9.9   & 2.3 \\
    Pet Supplies & 1.736 & 1.618 & 1.649 & 1.646 & 1.447 & 1.346 & 1.337 & \textbf{1.328} & 9.0   & 1.4   & 0.7 \\
    Office Products & 1.143 & 0.998 & 1.122 & 1.069 & 0.909 & 0.840 & 0.805 & \textbf{0.779} & 16.7  & 7.8   & 3.3 \\
    Android Apps & 1.922 & 1.871 & 1.805 & 1.842 & 1.517 & 1.515 & 1.509 & \textbf{1.494} & 1.5   & 1.4   & 1.0 \\
    Beauty & 1.950 & 1.711 & 1.631 & 1.552 & 1.453 & 1.404 & 1.409 & \textbf{1.387} & 4.8   & 1.2   & 1.6 \\
    Tools / Home & 1.569 & 1.310 & 1.356 & 1.314 & 1.208 & 1.122 & 1.101 & \textbf{1.096} & 10.2  & 2.4   & 0.5 \\
    Video Games & 1.627 & 1.665 & 1.576 & 1.568 & 1.307 & 1.276 & 1.269 & \textbf{1.257} & 4.0   & 1.5   & 1.0 \\
    Toys / Games  & 1.802 & 1.195 & 1.286 & 1.222 & 1.057 & 0.974 & 0.982 & \textbf{0.973} & 8.6   & 0.1   & 0.9 \\
    Health & 1.882 & 1.506 & 1.522 & 1.415 & 1.299 & 1.249 & 1.269 & \textbf{1.238} & 4.9   & 0.9   & 2.5 \\
    CellPhone & 1.972 & 1.668 & 1.622 & 1.606 & 1.524 & 1.431 & 1.452 & \textbf{1.413} & 7.9   & 1.3   & 2.8 \\
    Sports / Outdoors& 1.388 & 1.195 & 1.120 & 1.299 & 1.039 & 0.994 & 0.990 & \textbf{0.980} & 6.0   & 1.4   & 1.0 \\
    Kindle Store & 1.533 & 1.217 & 1.231 & 1.398 & 0.823 & 0.797 & 0.813 & \textbf{0.775} & 6.2   & 2.8   & 4.9 \\
    Home / Care & 1.667 & 1.547 & 1.584 & 1.654 & 1.259 & 1.235 & 1.237 & \textbf{1.220} & 3.2   & 1.2   & 1.4 \\
    Clothing & 2.396 & 1.492 & 1.462 & 1.535 & 1.322 & 1.197 & 1.207 & \textbf{1.187} & 11.4  & 0.8   & 1.7 \\
    CDs / Vinyl  & 1.368 & 1.555 & 1.432 & 1.368 & 1.045 & 1.010 & 1.018 & \textbf{1.005} & 4.0   & 0.5   & 1.3 \\
    Movies / TV & 1.690 & 1.852 & 1.518 & 1.775 & 1.960 & 1.176 & 1.187 & \textbf{1.144} & 71.3  & 2.8   & 3.8 \\
    Electronics & 1.962 & 2.120 & 1.950 & 1.651 & 1.372 & 1.365 & 1.368 & \textbf{1.350} & 1.6   & 1.1   & 1.3 \\
    \hline
    \end{tabular}%
     \caption{Performance comparison (mean squared error) on 24 benchmark datasets. The best performance is in boldface. $\Delta_{DC}, \Delta_{TN}, \Delta_{DA}$ are the relative improvements ($\%$) of MPCN over DeepCoNN (D-CON), TransNet (T-NET) and D-ATT respectively. MPCN achieves state-of-the-art performance, outperforming all existing methods on 24 benchmark datasets. }
  \label{tab:all_results}%
\end{table*}%

\subsection{Compared Methods}
We compare against a series of competitive baselines.
\begin{itemize}
\item \textbf{Matrix Factorization} (MF) is a standard and well-known baseline for CF. It represents the user and item rating with the inner product, i.e., $p^{\top}q$.
\item \textbf{Factorization Machines} (FM) \cite{DBLP:conf/icdm/Rendle10} are general purpose machine learning algorithms that use factorized parameters to model pairwise interaction within a real-valued feature vector. We concatenate the user-item latent embedding together and pass it through the FM model.
\item \textbf{Multi-layered Perceptrons} (MLP) are strong neural baselines for CF and were used as a baseline in \cite{He:2017:NCF:3038912.3052569}. We use the same pyramidal scheme of 3 layers.
\item \textbf{Neural Matrix Factorization} (NeuMF) \cite{He:2017:NCF:3038912.3052569} is the state-of-the-art model for interaction-only CF. It casts the MF model within a neural framework and combines the output with multi-layered perceptrons.
\item \textbf{Deep Co-Operative Neural Networks} (DeepCoNN) \cite{DBLP:conf/wsdm/ZhengNY17} is a review-based convolutional recommendation model. It trains convolutional representations of user and item and passes the concatenated embedding into a FM model.
\item \textbf{TransNet} \cite{catherine2017transnets} is an improved adaptation of DeepCoNN which incorporates transform layers and an additional training step that enforces the transformed representation to be similar to the embedding of the actual target review.
\item \textbf{Dual Attention CNN Model} (D-ATT) \cite{Seo:2017:ICN:3109859.3109890} is a recently proposed state-of-the-art CNN-based model that uses reviews for recommendation. This model is characterized by its usage of two forms of attentions (local and global). A final user (item) representation is learned by concatenating representations learned from both local and global attentions. The dot product between user and item representations is then used to estimate the rating score.
\end{itemize}
Given our already extensive comparisons against the state-of-the-art models, we omit comparisons with models such as HFT \cite{mcauley2013hidden}, Collaborative Topic Regression \cite{wang2011collaborative}, Collaborative Deep Learning (CDL) \cite{wang2015collaborative} and ConvMF \cite{Kim:2016:CMF:2959100.2959165} since they have been outperformed by the recently proposed DeepCoNN or D-ATT model.
\subsection{Experimental Setup}
The evaluation metric is the well-known (and standard) mean-squared error which measures the square error between the rating prediction and ground truth. We implement all models ourselves in Tensorflow. All models are trained with Adam \cite{DBLP:journals/corr/KingmaB14} with an initial learning rate of $10^{-3}$. We train all models for a maximum of $20$ epochs with early stopping (i.e., if model performance does not improve for 5 epochs) and report the test result from the best performing model on the development set. We found that models tend to converge before $20$ epochs. However, an exception is that the MF baseline requires many more epochs to converge. As such, we train the MF model till convergence. For interaction only models, the embedding size is set to $50$. For TransNet and DeepCoNN, the number of filters is set to $50$ and the filter size is $3$. For D-ATT, the global attention layer uses filter sizes of $[2,3,4]$. The word embedding layer is also set to $50$ dimensions. We regularize models with a dropout rate of $0.2$ and a fixed L2 regularization of $10^{-6}$. Dropout is applied after all fully-connected and convolutional layers. We use two transform layers in the TransNet model. All word embeddings are learned from scratch as we found that using pretrained embeddings consistently degrades performance across all datasets (and models). The maximum document length is set to $600$ words ($20$ reviews of $30$ tokens each) which we empirically found to be a reasonable length-specific performance bound. We assign a special delimiter token to separate reviews within a user (item) document for DeepCoNN, TransNet and D-ATT. If FM is used, the number of factors is set to $10$. For our proposed model, the number of pointers $p$ is tuned amongst $\{1,3,5,8,10\}$. On most datasets, the optimal performance is reached with $2-3$ pointers.

\subsection{Experimental Results}
Table \ref{tab:all_results} reports the results of our experiments. Firstly, we observe that our proposed MPCN is the top performing model on all 24 benchmark datasets. This ascertains the effectiveness of our proposed model and clearly answers \textbf{RQ1}. MPCN consistently and significantly outperforms DeepCoNN, TransNet and D-ATT, which are all recent competitive review-based methods for recommendation. The relative improvement is also encouraging with gains of up to $71\%$ (DeepCoNN), $19\%$ (TransNet) and $5\%$ (D-ATT). On majority of the datasets, performance gains are modest, seeing an improvement of $1\%-3\%$ for most models. Notably, the average percentage improvement of MPCN over DeepCoNN is $16\%$. The average performance gain over TransNet and D-ATT is a modest $3.2\%$ and $2.2\%$ respectively.

Pertaining to the relative ranking of the review-based models, our empirical evaluation reaffirms the claim of \cite{catherine2017transnets}, showing that TransNet always outperforms DeepCoNN. However, the relative ranking of D-ATT and TransNet switches positions frequently. Notably, TransNet uses the test review(s) as an additional data source (albeit as a training target) while D-ATT does not make use of this information. The additional training step of TransNet is actually quite effective and hypothetically could be used to enhance D-ATT or MPCN as well. However, we leave that for future work.

Next, the performance of interaction-only models (MF, FM, etc.) is consistently lower than review-based models (e.g., DeepCoNN). The relative performance of all interaction models is generally quite inconsistent over various datasets. However, one consistent fact is that MF performs worse than other models most of the time. The top scoring interaction model often switches between FM and MLP.

Finally, we give readers a sense of computational runtime. We provide an estimate that we found generally quite universal across multiple datasets. Let $t$ be the runtime of DeepCoNN, the runtime of MPCN $p=1$ is approximately $\approx 0.4t$. MPCN with $2$ and $3$ pointers are $0.8t$ and $1.2t$ respectively. TransNet and D-ATT run at $\approx 2t$. On medium size datasets (e.g., Amazon Beauty), $t \approx 40s$ on a GTX1060 GPU (batch size is $128$). We found that if $p_{opt}\leq2$, then MPCN is faster than DeepCoNN. While $p_{opt}\leq 5$ is the threshold for being equal with D-ATT and TransNet in terms of runtime.

\section{Hyperparameter \& Ablation Analysis}
In this section, we study the impact of key hyperparameter settings and various architectural choices on model performance.

\subsection{Ablation Analysis}
We study the impacts of various architectural decisions on model performance (\textbf{RQ2}). Table \ref{tab:ablation} reports an ablation analysis conducted on the
development sets of four benchmark datasets (\textit{Beauty, Office, Musical Instruments (M-Instr) and Amazon Instant Video (Inst-Vid)}). We report the results of several different model variations. We first begin describing the default setting in which ablation reports are deviated from. In the default setting, we use the standard model
with all components (review gates, word-level co-attention and FM prediction layer). The Multi-Pointer aggregation (aggr) is set to use a neural network (single layer nonlinear transform). The number of layers in the co-attention layer is set to $l=1$.

We report the validation results from 8 different variations, with the aims of clearly showcasing the importance of each component. In (1),
 we remove the review gating mechanism. In (2), we replace the FM with the simple inner product. In (3-4), we investigate the effects of different
 pointer aggregation (aggr) functions. They are the concatenate and additive operators respectively. In (5-6), we set $l=0$ (remove layer) and $l=2$. In (7), we remove the word level co-attention layer. In this case, the representation for user and item is simply the pointed review embedding. In (8), we remove the review-level co-attention (RLCA). This variation is no longer `hierarchical', and simply applies word-level co-attention to user and item reviews.
\begin{table}[H]
  \centering
    \begin{tabular}{|l|cccc|}
    \hline
    Architecture& \multicolumn{1}{c}{Beauty} & \multicolumn{1}{c}{Office} & \multicolumn{1}{c}{M-Instr} & \multicolumn{1}{c|}{Inst-Vid} \\
    \hline
    (0) Default & 1.290 & 0.770 & \textbf{0.827} & 0.975 \\
    (1) Remove Gates & 1.299 & \textbf{0.760} & 0.837 & 0.979 \\
    (2) Remove FM & 1.286 & 0.808 & 0.924 & 0.997 \\
    (3) Agrr (Concat) & \textbf{1.279} & 0.788 & 0.832 & \textbf{0.971} \\
    (4) Aggr (Additive) & 1.290 & 0.767 & 0.829 & 0.971 \\
    (5) set $l=0$ & 1.293 & 0.776 & 0.830 & 0.976 \\
    (6) set $l=2$ & 1.295 & 0.775 & 0.829 & 0.974 \\
    (7) Remove WLCA & 1.296 & 0.778 & 0.831 & 0.999 \\
    (8) Remove RLCA & 1.304 & 0.789 & 0.839& 0.1003 \\
    \hline
    \end{tabular}%
      \caption{Ablation analysis (validation MSE) on four datasets.}
  \label{tab:ablation}%
\end{table}%

Firstly, we observe the default setting is not universally the best across four datasets. As mentioned, the review gating mechanism helps in 3 out of 4 presented datasets. In the \textit{Office} dataset, removing the review gating layer improves performance. We found this to be true across most datasets, i.e., the review gating mechanism helps more often than not, but not always. The impacts of removing FM is quite easily noticed, leading to huge performance degradation on the \textit{M-Instr} dataset. Deprovement on \textit{Inst-Vid} and \textit{Office} is also significant. On the other hand, removing FM marginally improved performance on \textit{Beauty}. We also discovered that there is no straightforward choice of \textit{aggr} functions. Notably, the relative ranking of all three variants (concat, additive and neural network) are always interchanging across different datasets. As such, they have to be tuned. We also noticed that the choice of $l=1$ is safe across most datasets, as increasing or decreasing would often lead to performance degradation. Finally, removing the WLCA and RLCA consistently lowered performance on all datasets, which ascertains the effectiveness of the WLCA layer. Notably, removing RLCA seems to hurt performance more, which signifies that modeling at a review-level is essential.
\subsection{Effect of Number of Pointers}
Table \ref{ptr} reports the effect of varying pointers on performance ($\textbf{RQ3})$. We use 4 datasets of varying sizes as an illustrative example (\textit{Patio, Automotive, Sports} and \textit{Video Games}). The datasets shown are sorted from smallest to largest in terms of number of interactions. Clearly, we observe that the optimal number of pointers varies across all datasets. We found this to be true for the remainder datasets that are not shown. This seems to be more of a domain-dependent setting since we were not able to find any correlation with dataset size.  For most datasets, the optimal pointers falls in the range of $1-3$. In exceptional cases (\textit{Video Games}), the optimal number of pointers was $5$.
\begin{table}[htbp]
  \centering
    \begin{tabular}{|c|cccc|}
    \hline
    \multicolumn{1}{|c}{Ptr} & \multicolumn{1}{|c}{Patio} & \multicolumn{1}{c}{Automotive} & \multicolumn{1}{c}{Sports} & \multicolumn{1}{c|}{Video Games} \\
    \hline
    1     & 0.992 & \textbf{0.842} & \textbf{0.926} & 1.885 \\
    2     & 0.980  & 0.855 & 0.933 & 1.178 \\
    3     & \textbf{0.975} & 0.843 & 0.938 & 1.194 \\
    4     & 0.991 & 0.844 & 0.938 & 1.189 \\
    5     & 0.991 & 0.844 & 0.935 & \textbf{1.121} \\
    \hline
    \end{tabular}%
      \caption{Validation MSE on various datasets when varying the number of pointers. The best result is in boldface. The optimal number of pointers is domain-dependent.}
  \label{tab:ptr}%
\end{table}%

\begin{table*}
\begin{tabular}{|p{6.8cm}|p{6.8cm}|}
\hline
\textbf{User Review} & \textbf{Item Review}\\
\hline
game is really beautiful! coolest rpg ever... & ..a gift for a friend who really loves rpg games. he really loved it!\\
\hline
.. game is completely turned based, so you have time to ponder
you actions.. & ..is a great and engaging puzzler game but wasn't too challenging\\
\hline
just love this little guy ... phone is reasonably easy to put in and take out& ..case really fits the 5s like a glove.. \\
\hline
is a nice charger..but after a few momths it wasn't charging...& it is clearly a used or refurbished battery.. \\
\hline
cocoa is a wonderful, rich tasting, not overly sweet product ..& used to eat the dark and milk chocolate, and then i tried this and can't explain how good they are \\
\hline
\end{tabular}
\caption{Excerpts from top matching User and Item reviews extracted by MPCN's pointer mechanism. }
\label{aspects}
\end{table*}
\section{In-Depth Model Analysis}
In this section, we present several insights pertaining to the inner workings of our proposed model. This aims to answer \textbf{RQ4}.
\subsection{What are the pointers pointing to?}
In this section, we list some observations by analyzing the behavior of the MPCN model.
Firstly, we observed that pointers react to aspects and product sub-categories. In many cases, we observed that the pointer mechanism tries to find the most relevant review written by the user for a given product. If the target item is a phone case, then our model tries to find a review written by the user which is directed towards \textit{another} phone case product. Intuitively, we believe that this is trying to solicit the user's preferences about a type of product. We provide some qualitative examples in Table \ref{aspects}. Consider the context of
video games, it finds that the user has written a review about \textit{rpg} (roleplaying games). At the same time,
it finds that the item review consists of a (positive) review of the item being a good rpg game. As a result,
it surfaces both reviews and concurrently points to both of them. This follows suit for the other examples, e.g., it
finds a matching clue of \textit{puzzle games} (turn-based) in the second example. The last example is taken from the \textit{Gourmet
Food} dataset. In this case, it discovers that the user likes cocoa, and concurrently finds out that the product in question has some
relevant information about chocolate.

\subsection{Behavior of Multi-Pointer Learning}
\begin{figure}[H]
    \centering
    \begin{subfigure}[t]{0.22\textwidth}
        \centering
        \includegraphics[width=1.0\textwidth]{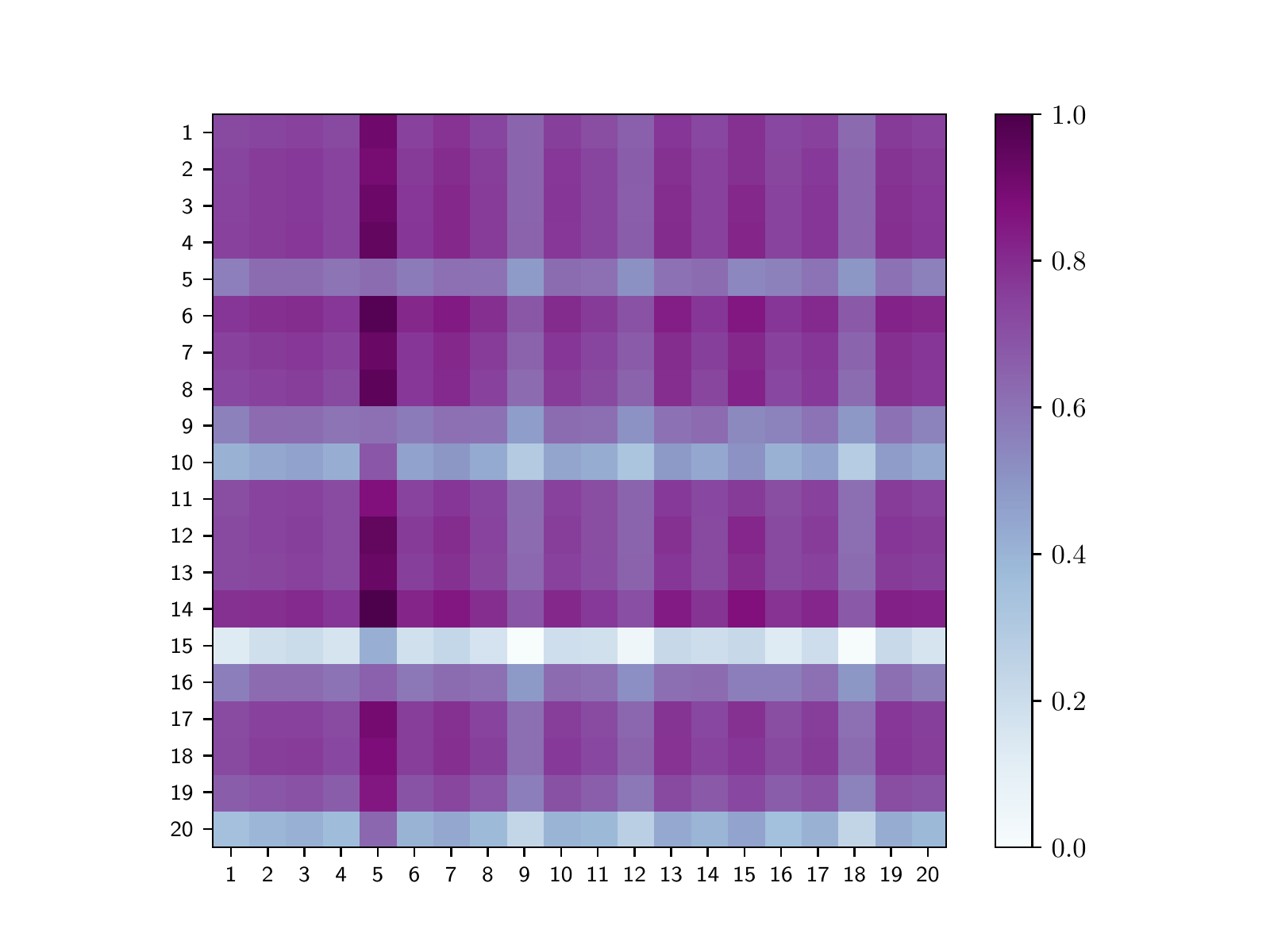}
        \caption{$p=1$}
    \end{subfigure}%
    \begin{subfigure}[t]{0.22\textwidth}
        \centering
        \includegraphics[width=1.0\textwidth]{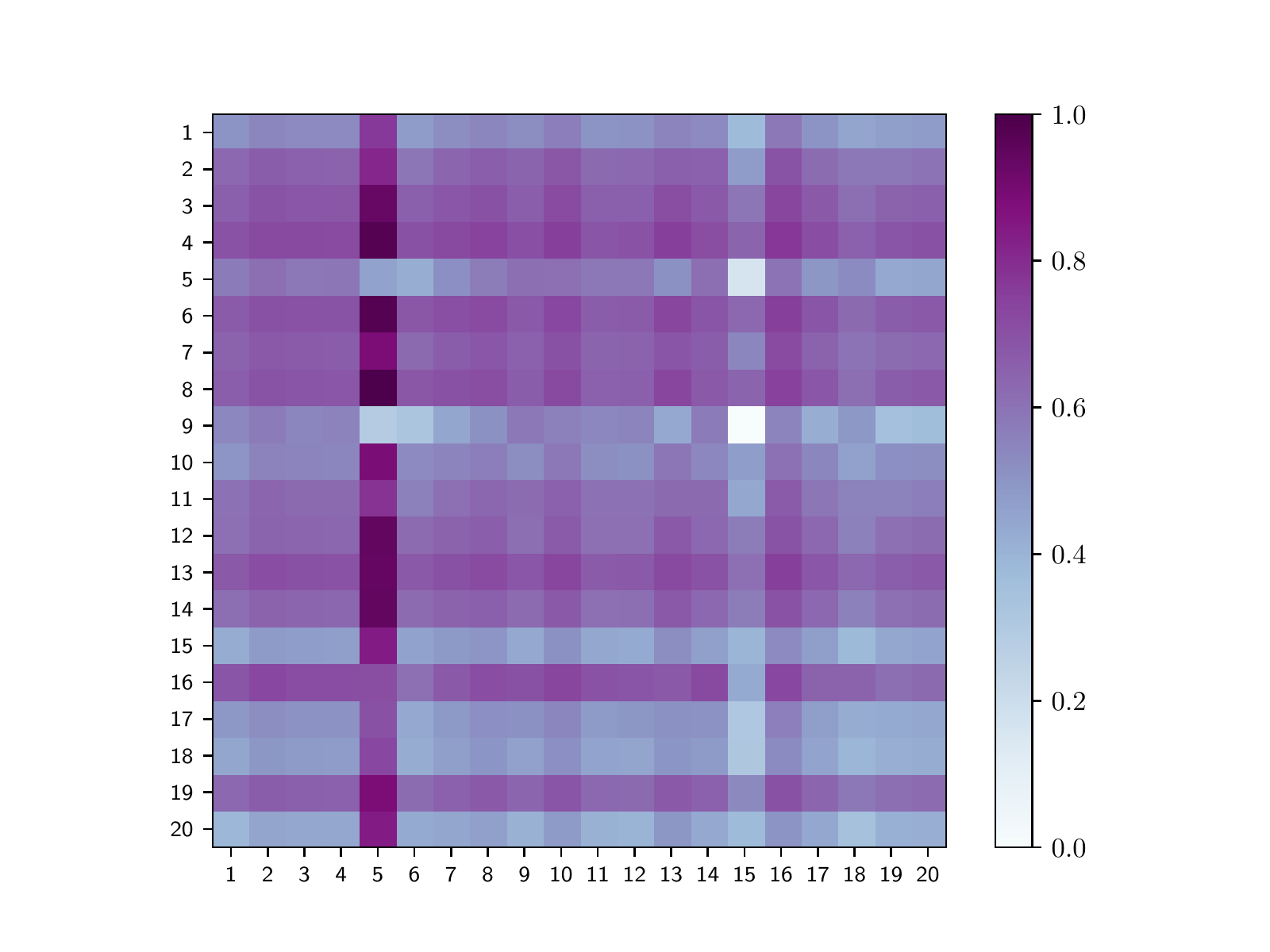}
        \caption{$p=2$}
    \end{subfigure}
      \begin{subfigure}[t]{0.22\textwidth}
        \centering
        \includegraphics[width=1.0\textwidth]{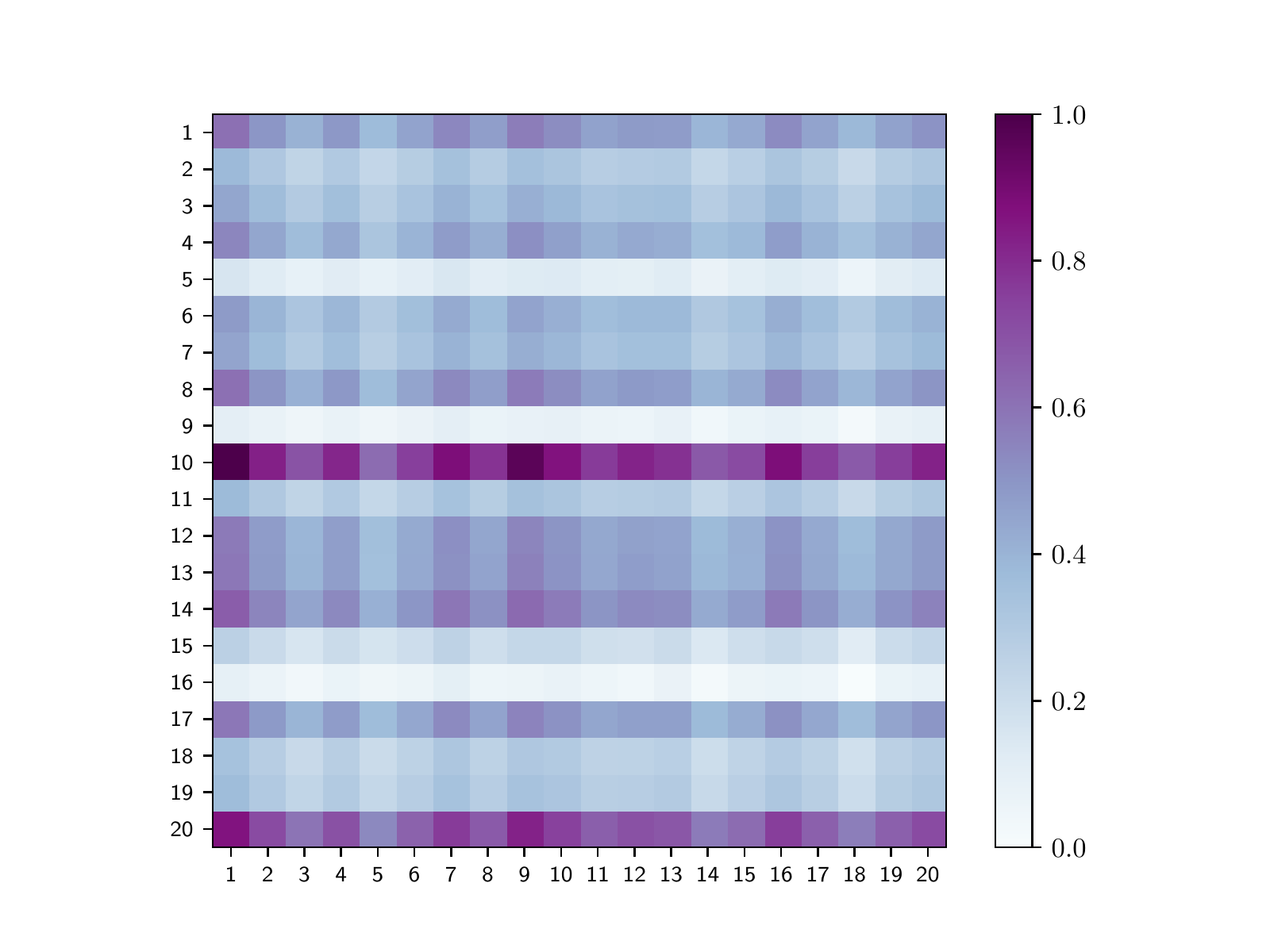}
        \caption{$p=3$}
    \end{subfigure}
   \begin{subfigure}[t]{0.22\textwidth}
        \centering
        \includegraphics[width=1.0\textwidth]{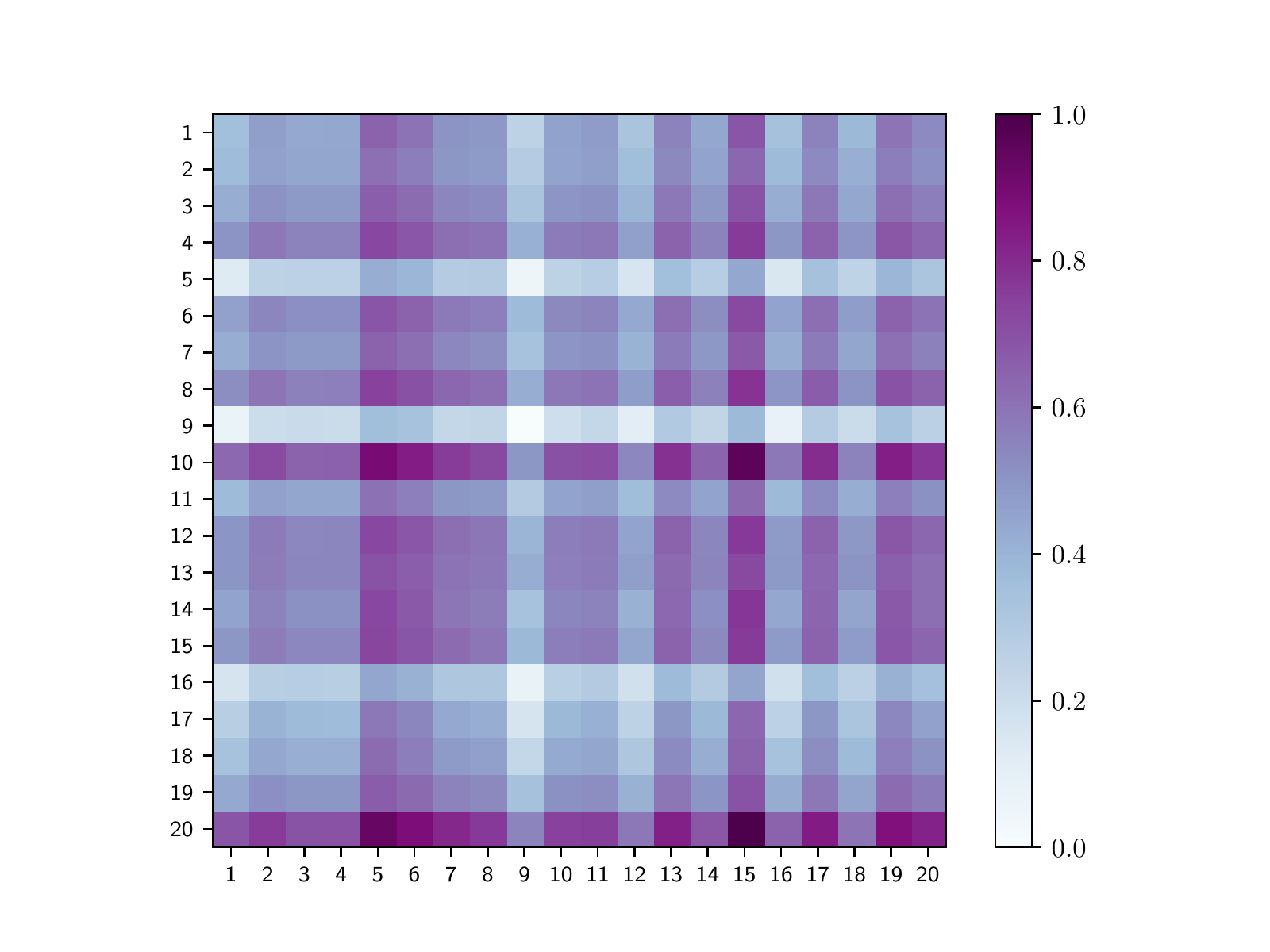}
        \caption{$p=4$}
    \end{subfigure}
    \caption{Visualisation of Review-level Co-Attention. Matching patterns
    differ significantly across multiple calls, hence generating different pointers.}
\label{reviewca}
\end{figure}

In this section, we study the behavior of our multi-pointer mechanism. First and foremost, this serves as another \textit{sanity check} and to observe if multi-pointers are really necessary, i.e., if pointers are not pointing all to the same reviews. Hence, this section aims to provide better insight into the inner workings of our proposed model. We trained a MPCN model with four pointers. A quick observation is that all four pointers point to different reviews (given the same user item pair). This is automatic, and does not require any special optimization constraint (such as explicitly enforcing the model to choose different reviews through an extra optimization term). Moreover, we analyze the affinity matrix from the review-level co-attention. Figure \ref{reviewca} shows the affinity matrix for pointers one to four.

\begin{table*}
  \centering
    \begin{tabular}{|c|ccccc|}
    \hline
    \multicolumn{1}{|c}{Condition} & \multicolumn{1}{|c}{D-Music} & \multicolumn{1}{c}{Apps} & \multicolumn{1}{c}{V-Games} & \multicolumn{1}{c}{Food} & \multicolumn{1}{c|}{Yelp17} \\
    \hline
    (1) All unique     & 85.2\% &  87.5\%  &  82.0\% & 99.2\% & 99.2\% \\
    (2) 1 Repeated & 13.2\% & 12.5\%  & 17.2\% & 0.7\% &0.7\% \\
    (3) All Repeated & 1.6\% &0.0\%  &1.6\% & 0.0\% & 0.0\%\\
    (4) One-to-Many & 64.8\% &57.8\% & 57.8\% &23.4\% & 43.7\%\\

    \hline
    \end{tabular}%
      \caption{Analysis of Multi-Pointer Behavior of MPCN on five datasets using $n_{p}=3$.}
  \label{tab:mpb}%
\end{table*}%
Secondly, it is also intuitive that it is not absolutely necessary for MPCN to always point to unique reviews given the same user-item pair. We observed a relatively significant \textit{one-to-many} pointer pattern on top of the usual \textit{one-to-one} pattern. In this case, the same review for user (item) is being matched with $n$ (many) different reviews from the item (user). This is also observed to be dataset / domain dependent. In a small minority of cases, all pointers pointed to the same reviews constantly (all repeated condition). However, this is understandable as there is just insufficient information in the user and item review bank. Additionally, we analyzed a small sample from the test set, determining if any of the following conditions hold for each test case.

Table \ref{tab:mpb} reports the percentages of test samples which falls into each category. We report results on five datasets \textit{Digital Music (D-Music)}, \textit{Android Apps (Apps)}, \textit{Video Games (V-Games)}, \textit{Gourmet Food (Food)} and \textit{Yelp17}. Here, we observe that pointer behavior is largely influenced by domain. The first three are concerned with electronic domains while the last two are more focused on food (and restaurants). We clearly observe that Food and Yelp have very similar pointer behavior. In general, the electronic domains usually make an inference using a fewer subsets of reviews. This is made evident by the high one-to-many ratio which signifies that there is often one important review written by the user (or item) that contributes more (and needs to be matched with multiple opposing reviews). Conversely, the food domains require more evidences across multiple reviews. We believe this is one of the biggest insights that our work offers, i.e., shedding light on how evidence aggregation works (and differs across domains) in review-based recommendation.

\section{Conclusion}
We proposed a new state-of-the-art neural model for recommendation with reviews. Our proposed Multi-Pointer Co-Attention Networks outperforms many strong competitors on 24 benchmark datasets from Amazon and Yelp. We conduct extensive analysis on the inner workings of our proposed multi-pointer learning mechanism. By analyzing the pointer behavior across multiple domains, we conclude that different domains (such as food-related and electronics-related) have different \textit{`evidence aggregation'} patterns. While our model dynamically handles this aspect, we believe this warrants further investigation.

\section{Acknowledgements}
The authors thank anonymous reviewers of KDD 2018 for their time and effort to review this paper.

\bibliographystyle{ACM-Reference-Format}
\balance
\bibliography{references} {}

\end{document}